\documentclass[10pt,journal,cspaper,compsoc]{IEEEtran}

\usepackage{booktabs} 
\usepackage{epsfig}
\usepackage{graphicx}
\usepackage{amsmath}
\usepackage{amssymb}
\usepackage{algorithm}
\usepackage{algorithmic}
\usepackage{subfigure}
\usepackage{url}
\usepackage{color}
\usepackage{multirow}
\usepackage[usenames,dvipsnames,table]{xcolor}
\usepackage{times}

\usepackage[pagebackref=true,breaklinks=true,letterpaper=true,colorlinks,bookmarks=false]{hyperref}

\begin{document}
	
	\title{Learning to Segment Human by Watching YouTube}
	
	\author{Xiaodan~Liang, Yunchao~Wei, Liang~Lin, Yunpeng~Chen, Xiaohui~Shen, Jianchao~Yang,  Shuicheng~Yan
		\IEEEcompsocitemizethanks{\IEEEcompsocthanksitem Xiaodan Liang and Liang Lin are with Sun Yat-sen University, China. (Corresponding author: Liang Lin) Email: xdliang328@gmail.com; linliang@ieee.org. Xiaohui Shen is with Adobe Research and  Jianchao Yang is with SnapChat, U.S. Email: xshen@adobe.com; jcyangenator@gmail.com. Yunchao Wei and Yunpeng Chen are with  National University of Singapore. Email: wychao1987@gmail.com; chenyunpeng@u.nus.edu. Shuicheng Yan is with 360 AI Institute and National University of Singapore. Email: eleyans@nus.edu.sg. }\protect\\
		
		\thanks{}}
	
	\markboth{IEEE TRANSACTIONS ON PATTERN ANALYSIS AND MACHINE INTELLIGENCE, VOL. 39, NO. 7, JULY 2017}
	{Shell \MakeLowercase{\textit{et al.}}: Bare Demo of IEEEtran.cls for Computer Society Journals}

	
	
	\IEEEcompsoctitleabstractindextext{%
		\begin{abstract}
			An intuition on human segmentation is that when a human is moving in a video, the video-context (e.g., appearance and motion clues) may potentially infer reasonable mask information for the whole human body.  Inspired by this, based on popular deep convolutional neural networks (CNN), we explore a very-weakly supervised learning framework for human segmentation task, where only an imperfect human detector is available along with massive weakly-labeled YouTube videos. In our solution, {the video-context guided human mask inference and CNN based segmentation network learning iterate to mutually enhance each other until no further improvement gains.} In the first step, each video is decomposed into supervoxels by the unsupervised video segmentation. The superpixels within the supervoxels are then classified as human or non-human by graph optimization with unary energies from the imperfect human detection results and the predicted confidence maps by the CNN trained in the previous iteration. In the second step, the video-context derived human masks are used as direct labels to train CNN. Extensive experiments on the challenging PASCAL VOC 2012 semantic segmentation benchmark demonstrate that the proposed framework has already achieved superior results than all previous weakly-supervised methods with object class or bounding box annotations. In addition, by augmenting with the annotated masks from PASCAL VOC 2012, our method reaches a new state-of-the-art performance on the human segmentation task.
		\end{abstract}
	}
	
	\maketitle
	
	\IEEEdisplaynotcompsoctitleabstractindextext
	
	\IEEEpeerreviewmaketitle


\section{Introduction}

Recently, tremendous advances in semantic segmentation have been made~\cite{BoxSup}~\cite{wcrf}~\cite{long2014fully}~\cite{CRF-RNN}. These approaches often rely on deep convolutional neural networks (CNN)~\cite{vgg} trained on a large-scale classification dataset~\cite{imagenet}, which is then transfered to the segmentation task based on the mask annotations~\cite{long2014fully}~\cite{wcrf}~\cite{BoxSup}. However, the annotation for pixel-wise segmentation masks usually requires considerable human effort. In addition, the construction of a semantic segmentation dataset covering diverse appearances, view-points or scales of objects is also costly and difficult. These limitations hinder the development of semantic segmentation  which generally requires large-scale data for training.

\begin{figure*}
	\begin{center}
		\includegraphics[scale=0.20]{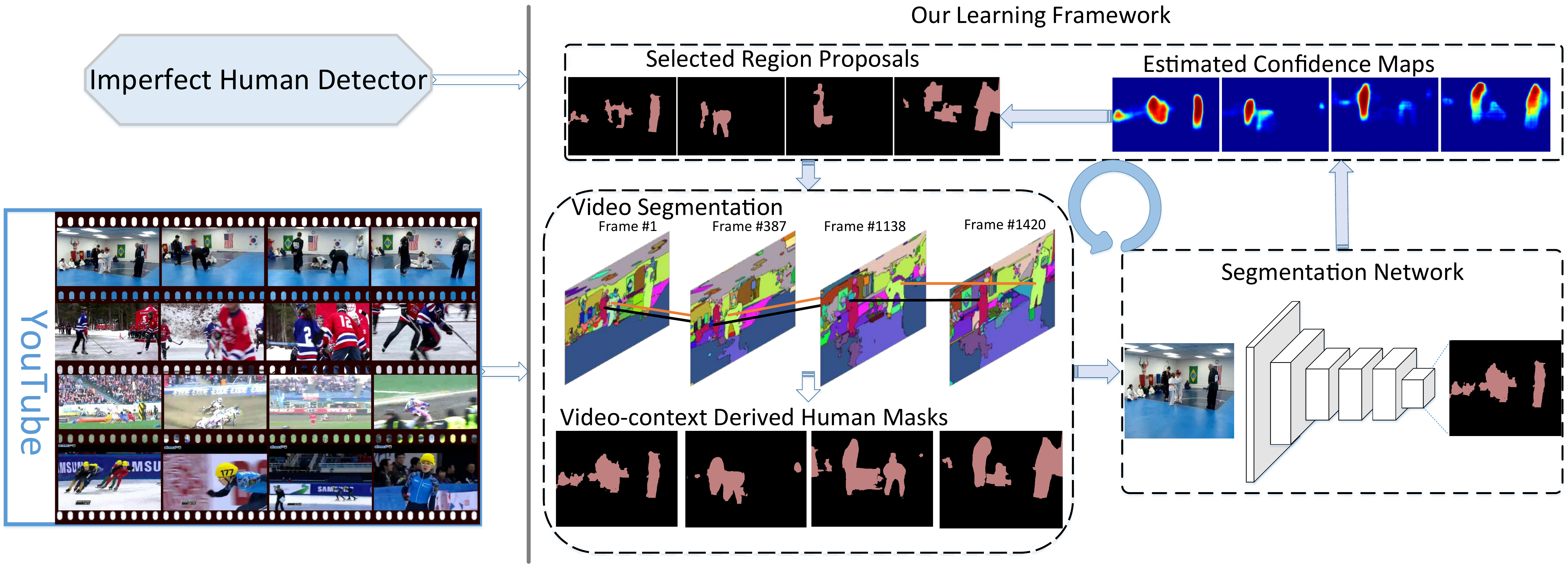}
		\vspace{-2mm}
		\caption{{An overview of our framework. Given raw YouTube videos, the proposed framework learns a good segmentation network by iteratively inferring video-context derived human masks and updating the network. For each video, we first extract mid-level supervoxels (unique colors represent unique supervoxels, and same colors across frames denote the same supervoxels) in the key frames. For each key frame, we use a pre-trained imperfect human detector to initialize the location of each instance of interest. The spatio-temporal graph optimization is then performed to extract video-context derived human masks, which are then utilized to update the segmentation network. As a feedback, the updated network will provide refined confidence maps to help generate better human masks.}}
		\vspace{-8mm}
		\label{fig:framework}
	\end{center}
\end{figure*}

{While a large collection of fully annotated images are difficult to obtain, weakly-labeled yet related videos are abundant on video sharing websites, e.g., YouTube.com, especially for human segmentation task. Intuitively, when a human instance is moving in the video, the inherent motion cues with the aid of an imperfect human detector may potentially help identify the human masks out of the background~\cite{jain2014supervoxel}\cite{tang2013discriminative}\cite{zhang2013video}.
Thus in this paper, we target at using the video-context derived human masks from raw YouTube videos to iteratively train and update a good segmentation neural network, instead of using a limited number of single image mask annotations like in traditional approaches.} {The video-context is used to infer the human masks by exploiting spatial and temporal contextual information over video frames.} Note that our framework can be applied to general object segmentation tasks, especially for moving objects. { This paper focuses on human segmentation, as human-centric videos are the most common on YouTube.} { Figure~\ref{fig:framework} provides an overview of our unified framework containing two integrated steps, i.e., the video-context guided human mask inference and the CNN-based human segmentation network learning.} 

In the first step, given a raw video, we extract the supervoxels, which are the spatio-temporal analogs of superpixels, to provide a bottom-up volumetric segmentation that tends to preserve object boundaries and motion continuousness~\cite{corsosupervoxel}. The spatio-temporal graph is built on the superpixels within supervoxels. To remove the ambiguity in determining the instances of interest in the video, we resort to an imperfect human detector~\cite{babylearning} and region proposal method~\cite{MCG} to generate the candidate segmentation masks. These masks are then combined with the confidence maps predicted by the currently trained CNN to provide the unary energy of each node. The graph-based optimization is then performed to find optimal human label assignments of the nodes by maximizing both appearance consistency within the neighboring nodes and the long-term label consistency within each supervoxel. 

In the second step, the video-context derived human masks extracted from massive raw YouTube videos are then utilized to train and update a deep convolutional neural network (CNN). {One important issue in training with those raw videos is the existence of noisy labels within these extracted masks. To effectively reduce the influence of noisy data, }we utilize the sample-weighted loss during the network optimization. 
The trained network in turn makes better segmentation predictions for the key frames in each video, which can help refine the video-context derived human masks. This process iterates to gradually update the video-context derived human masks and the network parameters until the network is mature. 

We evaluate our method on the PASCAL VOC 2012 segmentation benchmark~\cite{voc}. Our very-weakly supervised learning framework by using raw YouTube videos achieves significantly better performance than the previous weakly supervised methods (i.e., using box annotations)~\cite{BoxSup}~\cite{wcrf} as well as the fully supervised (i.e., using mask annotations) methods~\cite{DeepLabCRF}~\cite{long2014fully}~\cite{CRF-RNN}. By combining with limited annotated data, our weakly supervised variant (i.e., using the box annotations on VOC) and the semi-supervised variant (i.e., using the mask annotations on VOC) yield superior accuracies than the previous methods~\cite{wcrf}~\cite{BoxSup} using extensive extra 123k annotations on Microsoft COCO~\cite{lin2014microsoft}. {Note that the general image-level supervision is also utilized in our approach as we pre-train our neural network on ImageNet.} 

\section{Related Work}

\textbf{Semantic Segmentation.} Deep convolutional neural networks (CNN) have achieved great success with the growing training data on object classification~\cite{szegedy2014going}~\cite{alexnet}~\cite{zhang2015bit}~\cite{chen2016disc}. However, current available datasets for object detection and segmentation often contain a relatively limited number of labeled samples. Most recent progress on object 
segmentation~\cite{long2014fully}~\cite{wcrf}~\cite{BoxSup} was achieved by fine-tuning the pre-trained classification network with limited mask annotations. These limited data hinder the advance of semantic segmentation to more challenging scenarios. Existing segmentation methods~\cite{wcrf}~\cite{BoxSup}~\cite{autoannotation} explored using bounding box annotations instead of mask annotations. Differ from these previous methods, our framework iteratively refines the video-context derived human masks using the updated segmentation network, and in turn improves the network based on these masks. Differ from the semi-supervised learning (using a portion of mask annotations) and weakly supervised learning (using the object class or bounding box annotations), the proposed very-weakly supervised learning only relies on weakly labeled videos and an imperfect human detector. It is true that training the human detector needs a certain number of bounding box annotations. {Instead of directly using those annotated boxes to train the human segmentation network, our model can be progressively improved by gradually mining more variant instances from weakly labeled videos.}
	

\textbf{Video Segmentation.} Unsupervised video segmentation focused on extracting coherent groups of supervoxels by considering the appearance and temporal consistency. 
These methods tend to over-segment an object into multiple parts and provide a mid-level space-time grouping, which cannot be directly used for object segmentation. Recent approaches proposed to upgrade the supervoxels to object-level segments~\cite{tang2013discriminative}~\cite{jain2014supervoxel}. 
Their performance is often limited by the incorrect segment masks.  
   
{\textbf{Semi-supervised Learning.} To minimize human efforts, some image-based attempts~\cite{NEILiccv13}~\cite{ShrivastavaSG12}~\cite{LEARN}~\cite{AddUnlabeled}~\cite{russakovsky2015best}~\cite{WangTCSVT2016} have been devoted to learning reliable models with very few labeled data for object detection. Among these methods, the semantic relationships~\cite{NEILiccv13} were further used to provide more constraints on selecting instances. In addition, the video-based approaches~\cite{weakannvideoDetect12}~\cite{wang2015unsupervised}~\cite{Xu_2015_CVPR}~\cite{joulin2014efficient} utilized motion cues and appearance correlations within video frames to augment the model training. 
}

\vspace{-3mm}
\section{Our Framework}

Our framework is
Figure~\ref{fig:framework} illustrates our very-weakly supervised learning framework for video-context guided human segmentation.  

\vspace{-2mm}
\subsection{The Iterative Learning Procedure}
\vspace{-1mm}

The proposed framework is applicable for training all fully supervised network structures based on CNN, such as FCN~\cite{long2014fully} and DeepLab-CRF method~\cite{DeepLabCRF}. In this paper, we adopt the original version of DeepLab-CRF method~\cite{DeepLabCRF} (i.e., without using multiscale and Large-FOV) as the basic structure due to its leading accuracy and competitive efficiency. Also, many weakly supervised competing methods~\cite{wcrf}~\cite{BoxSup}  using object class annotations and bounding box annotations only reported their results based on DeepLab-CRF~\cite{DeepLabCRF}. 

Our learning process is iteratively performed to train and update the network with the video-context derived human masks and then refine these masks based on the improved network. {Note that the category-level human annotations on ImageNet are used since our segmentation model is finetuned on the pre-trained VGG model. The human masks generated from YouTube videos may be labeled with incorrect categories.} These noises may degrade the performance of our framework, especially in the early iterations where the learned network is more vulnerable to noises. {To reduce disturbance of noisy labels, the sample-weighted loss is utilized to train the network.} During the network training, more considerations should be given to the video-context derived human masks with higher labeling quality\footnote{The higher labeling quality means that a derived human mask contains fewer incorrectly labeled pixels, as defined in Section~\ref{sec:videoSegmentation}.}. Suppose in the $t$-th iteration of the learning process, we collect $\{I_i\}_{i=1}^N$ training frames from the videos. For each training frame $I_i$ selected from the video set $\mathbf{v}$, the video-context derived human mask $l^{t-1}_{i}$ is inferred by conditioning on the network parameter $\theta^{t-1}$ in the $(t-1)$-th iteration and the video information, i.e., $l^{t-1}_{i} = \varphi(I_i, \theta^{t-1}, \mathbf{v})$. The corresponding labeling quality for $l^{t-1}_{i}$ is denoted as $\omega^{t-1}_i$. {The network optimization in every iteration is thus formulated as a pixel-level regression problem  from the training images to the generated masks.} Specifically, the objective function to be minimized can be written as

\vspace{-6mm}
\begin{equation}
\small
 \begin{split} 
L(\theta^{t}) &= \frac{1}{N}\sum_{i=1}^{N} \omega_i^{t-1} \cdot \frac{1}{M_i}\sum_{j}^{M_i} e(P_{i,j}(\theta^t), l^{t-1}_{i,j})\\
            &= \frac{1}{N}\sum_{i=1}^{N} \omega_i^{t-1} \cdot \frac{1}{M_i}\sum_{j}^{M_i} e(P_{i,j}(\theta^t), \varphi_j(I_i, \theta^{t-1}, \mathbf{v})), 
\end{split}
\end{equation} 
\vspace{-4mm}

\noindent{where} $l^{t-1}_{i,j}$ is the target label at the $j$-th pixel of the image $I_i$, $M_i$ indicates the pixel number of each image $I_i$ and $P_{i,j}(\theta^t)$ is the predicted pixel-level label produced by the network with the parameter $\theta^{t}$. $e(P_{i,j}(\theta^t), l^{t-1}_{i,j})$ is the pixel-wise softmax loss function.  The targets to be optimized in our task are the network parameters and the video-context derived human masks of all training frames. 

An iterative learning procedure is proposed to find the solution. With the video-context derived human mask $l^{t-1}_i$ and labeling quality $\omega^{t-1}_i$ for each training frame fixed, we can update the network parameter $\theta^{t}$. The problem thus becomes a segmentation network learning problem with the sample-weighted loss. The parameter $\theta^{t}$ can be updated by back-propagation and stochastic gradient descent (SGD), as in~\cite{DeepLabCRF}. In turn, after the network parameter $\theta^{t}$ in the $t$-th iteration is updated, we can refine the video-context derived human masks $l^{t}_i$ by the video-context guided inference, i.e., inferring the $\varphi(I_i, \theta^{t}, \mathbf{v})$. We will give more details about the computation of $\varphi(\cdot)$ and $\omega^{t-1}_i$ in Section~\ref{sec:videoSegmentation}. The segmentation network is initialized by the model pre-trained on the ImageNet classification dataset. {In this way, the general image-level supervision provided by the ImageNet is naturally embedded in our models}. This network is then trained in every iteration based on the refined video-context derived human masks. {This process is iteratively performed until no further improvement is observed. The fully-connected conditional random field (CRF)~\cite{DeepLabCRF} is adopted to further refine the results.} 

\vspace{-2mm}
\subsection{Video-context Guided Human Mask Inference}

\label{sec:videoSegmentation}
In this subsection, we introduce the details of generating the video-context derived human masks from raw videos along with an imperfect human detector. Given the network parameter $\theta^t$, the video-context derived human masks are obtained by solving $l_i^t = \varphi(I_i, \theta^{t}, \mathbf{v})$, and the labeling quality $\omega^t_i$ for each frame $I_i$ is accordingly predicted. The index $i$ is omitted for simplicity in the following.  We crawl about 5,000 videos  which may contain humans from \emph{YouTube.com}. {Following~\cite{babylearning}, we use the keywords from the PASCAL VOC collection to prune the videos that are unrelated to the person category. In the crawled videos, the collected video set includes approximate $30\%$ noisy videos that contain none of person instances.} The spatio-temporal graph optimization is performed to extract video-context derived human masks. 

\textbf{Video Pre-processing.} For each video, the supervoxels are first extracted, which are space-time regions computed with the mid-level unsupervised video segmentation method~\cite{li2016}. We empirically extract all the supervoxels at the 10-th level of the tree, which is a good tradeoff between semantic meaning and fine detail preserving properties. Though the supervoxels for each instance are unlikely to remain stable through the whole video due to pose changing and background cluttering, the supervoxels often persist for a series of frames due to the temporal continuity. Each video can be split into many video shots which are divided when over half of the supervoxels across two adjacent frames change (e.g., the supervoxels of some objects are lost or a new object appears). 

\textbf{Spatio-temporal Graph Optimization.} We project each supervoxel into each of its children frames to obtain the corresponding spatial superpixel as a node of the graph. 
Notably, the object boundary within each supervoxel can be better preserved in the key frame where large appearance and motion changes occur. {We thus select the key frames as the initial candidate set by the criterion that more than $10$ supervoxels change compared with their previous frame.} The graph optimization is performed on these selected key frames in each shot. 

\begin{figure}
	\begin{center}
		\includegraphics[scale=0.16]{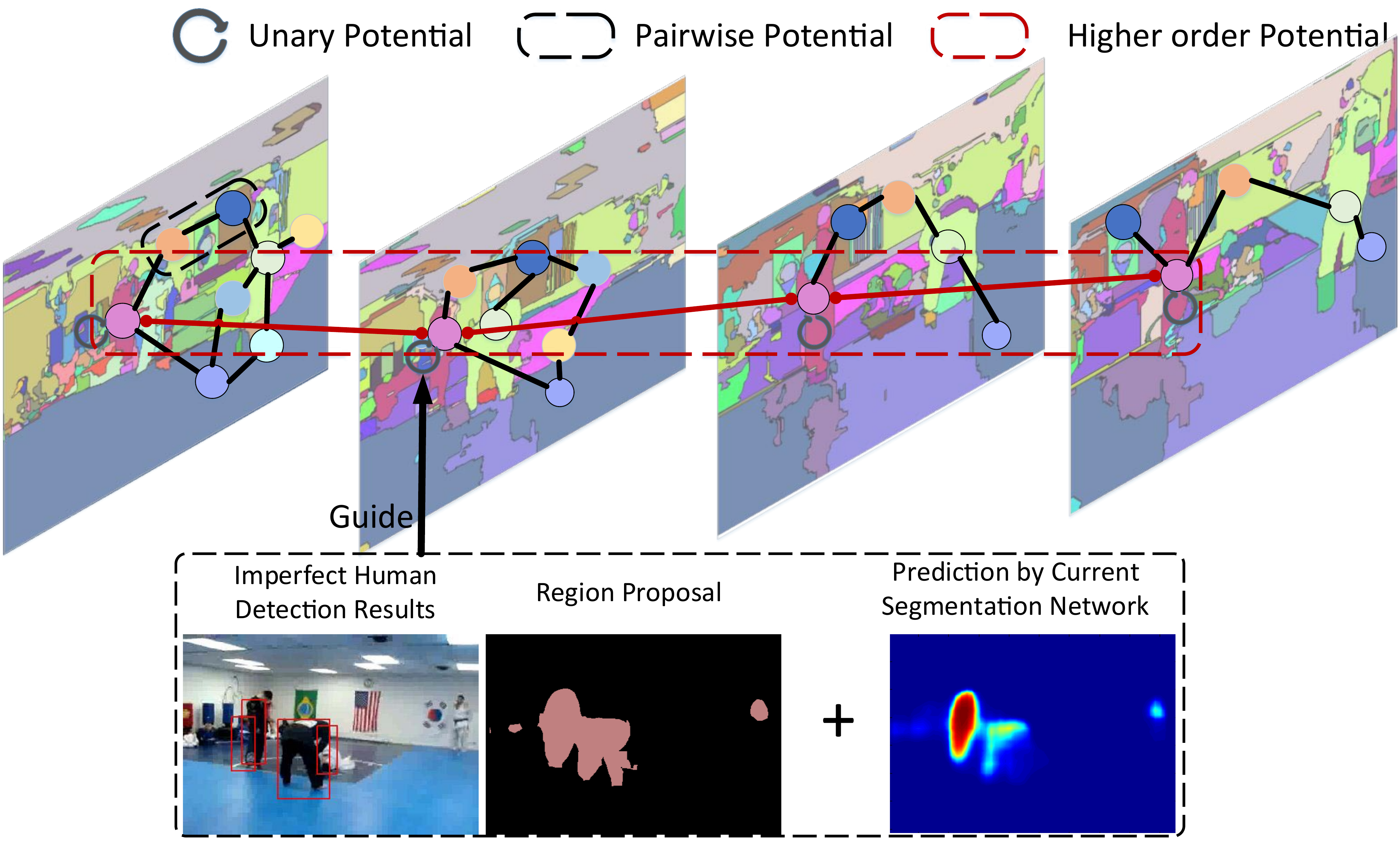}
		\vspace{-1mm}
		\caption{{{Illustration of the spatio-temporal graph. Nodes are spatial superpixels in every key frame. For legibility, only a small subset of nodes and connections are depicted. Best viewed in color.}}}
		\vspace{-6mm}
		\label{fig:graph}
	\end{center}
\end{figure}

Formally, the spatio-temporal graph structure $G$ consists of the nodes $\mathcal{B}$ and the edges $\mathcal{E}$, as shown in Figure~\ref{fig:graph}. Let $\mathcal{B} = \{B_k\}_{k=1}^K$ be the set of spatial superpixel nodes over the entire video, where $K$ refers to the number of key frames. $B_k$ contains nodes belonging to the $k$-th frame, which is a collection of nodes $\{b_k^c\}_{c=1}^{C_k}$, where $C_k$ is the number of nodes in the $k$-th frame.  We assign the variable $y_k^c \in \{+1, -1\}$ to each node, which is either human (+1) or other content (-1). The target is to obtain a labeling $\mathcal{Y} = \{Y_k\}_{k=1}^K$, where $Y_k = \{y_k^c\}_{c=1}^{C_k}$ denotes the labels of nodes belonging to the $k$-th frame. The edge set $\mathcal{E}$ is defined as the set of spatial edges. A spatial edge exists between the neighboring pair of nodes $(b_k^c, b_k^{c'})$ in a given key frame. Finally, we use $\mathcal{S}$ to denote the set of supervoxels. Each element $s\in \mathcal{S}$ represents each supervoxel. We denote $y_s$ as the set of labels assigned to the nodes within the supervoxel $s$. For each node $b_k^c$, we compute its visual feature, i.e, the concatenation of bag-of-words (75 dim) from RGB, Lab and HOG descriptors. The visual dissimilarity between two nodes $D(b_k^c, b_k^{c'})$ is computed by the Euclidean distance.

To enforce the local label smoothness and long-range temporal coherence over the supervoxels, the energy function over $G = (\mathcal{B}, \mathcal{E})$ is defined as

\vspace{-4mm}
\begin{equation}
\small
 \begin{split}
E(\mathcal{Y}) &= \underbrace{\sum_{(k,c)} \Phi_k^c(y_k^c)}_{Unary} + \underbrace{\sum_{[(k,c), (k, c')]\in \mathcal{E}} \Phi_k^{c,c'}(y_k^c, y_k^{c'})}_{Pairwise}\\
 & + \underbrace{\sum_{s\in \mathcal{S}}\Phi_s(y_s)}_{Higher\ order}.
\end{split}
\label{enegy}
\end{equation}
\vspace{-3mm}

The optimal human masks are obtained by minimizing  Eqn.~(\ref{enegy}): $\mathcal{Y}^* = \text{argmin}_{\mathcal{Y}}E(\mathcal{Y})$. The unary potential $\Phi_k^c(y_k^c)$ accounts for the cost of assigning each node as the human or others. The pairwise potential $\Phi_k^{c,c'}(y_k^c, y_k^{c'})$ promotes smooth segmentation by penalizing neighboring nodes with different labels. The higher order potential $\Phi_s(y_s)$ ensures long term label consistency along each supervoxel.

\emph{Unary Potential:} The unary potential of each node is computed based on the imperfect human detection results and the predicted confidence maps by the updated CNN in the previous iteration. First, to roughly locate the instance of interest, human detection is performed on each key frame $I_k$. We use the object detection method in~\cite{babylearning} to detect human instances and only those boxes with scores higher than $-1$ are selected. For each detected box, an optimal region proposal is selected. The state-of-the-art region proposal method, i.e., Multiscale Combinatorial Grouping (MCG)~\cite{MCG}, is adopted to generate about 2,000 region proposals per image. Denote $\mathbf{r}$ and $\mathbf{h}$ as the candidate proposal and the detected box, respectively. For each box $\mathbf{h}$, we expect to pick out a candidate proposal $\hat{\mathbf{r}}$ that has a large overlap with the box and also a high estimated confidence from the CNN prediction $\hat{\mathbf{g}}_k^t$. The $\hat{\mathbf{r}}$ is computed by
   
\vspace{-2mm}
\begin{equation}
\small
	 \hat{\mathbf{r}} = \text{argmax}_{\mathbf{r}} (\text{IoU}(\mathbf{r} ,\mathbf{h}) + \frac{1}{|\mathbf{r}|}\sum\hat{\mathbf{g}}_k^t(\mathbf{r})),
	 \label{eq:region}
\end{equation} 
\noindent{where} $\text{IoU}(\mathbf{r} ,\mathbf{h}) \in [0,1]$ is the intersection-over-union ratio computed from the box $\mathbf{h}$ and the tight bounding box of the region proposal $\mathbf{r}$. The term $\frac{1}{|\mathbf{r}|}\sum\hat{\mathbf{g}}_k^t(\mathbf{r})$ denotes the mean of the predicted confidences within region proposal $\mathbf{r}$. {The proposals with the same tight bounding box can be distinguished by their predicted confidences.} The optimal region proposal $\hat{\mathbf{r}}$ for each detected box can be gradually refined along with the prediction $\hat{\mathbf{g}}_k^t$ by the improved CNN. We use ${R}_k$ to denote the generated proposal mask for each frame $I_k$ by combining all selected proposals $\hat{\mathbf{r}}$. The probability of each node $b_k^c$ to be the human is computed as

\vspace{-6mm}
\begin{equation}
\small
P(y_k^c = 1) = \lambda_r\eta(y_k^c, {R}_k) + \lambda_e\frac{1}{|y_k^c|} \sum\hat{\mathbf{g}}_k^t(y_k^c),
\label{eq:unary}
\end{equation} 
\vspace{-4mm}

\noindent{where} $\lambda_r$ and $\lambda_e$ are set as $0.5$ empirically. $\eta(y_k^c, {R}_k)$ represents the percentage of the spatial super-pixel node $b_k^c$ contained within the mask ${R}_k$. $|y_k^c|$ is the pixel count within the node $b_k^c$, and $\sum\hat{\mathbf{g}}_k^t(y_k^c)$ is the sum of the predicted probabilities. The unary potential of each node is computed by

\vspace{-4mm}
\begin{equation}
\small
 \begin{split}
\Phi_k^c(y_k^c) = \left\{
\begin{array}{lcl}
-\log(P(y_k^c = 1)) & \text{if} \ y_k^c = +1, \\
-\log(1 - P(y_k^c = 1)) & \text{if} \  y_k^c = -1.
\end{array}  
\right.
\end{split}
\label{eq:unary}
\end{equation} 

 After the network is updated, the unary potential can be accordingly updated to generate better video-context derived human masks. Note that in the beginning of our learning process, the $\sum\hat{\mathbf{g}}_k^t(y_k^c)$ part will be eliminated in Eqn.~(\ref{eq:region}) and~(\ref{eq:unary}) to infer the human masks, since the segmentation network is not yet trained. { Based on the initialized bounding box generated by the imperfect object detector, our method can gradually segment all human masks in the video.}

\emph{Pairwise Potential:} We use the standard pairwise terms for edges to ensure the local label smoothness:

\vspace{-4mm}
\begin{equation}
\small
\begin{split}
\Phi_k^{c,c'}(y_k^c, y_k^{c'}) = \delta(y_k^c \neq y_k^{c'}) \exp(-\beta_p D(b_k^c, b_k^{c'})),
\end{split}
\end{equation} 
\vspace{-4mm}

\noindent{where} $\beta_p$ is set as the inverse of the mean of all individual distances, following~\cite{jain2014supervoxel}. 

\emph{Higher Order Potential:} The supervoxel label consistency potential is defined to ensure the long-term coherence within each supervoxel. We adopt the Robust $P^n$ model~\cite{kohli2009robust} to define this potential:

\vspace{-4mm}
\begin{equation}
\small
\Phi_s(y_s) = \left\{
\begin{array}{lcl}
{N(y_s)\frac{1}{Q}\lambda_{max}(s)} &\text{if} \ N(y_s) \leq Q, \\
{\lambda_{max}(s)} & \text{otherwise},
\end{array}  
\right.
\end{equation}
\vspace{-4mm}

\noindent{where} $y_s$ denotes the labels of all nodes within the supervoxel $s\in \mathcal{S}$, and $N(y_s)$ is the number of nodes within the supervoxel $s$ that is not assigned with the dominant label, i.e., $N(y_s) = \min(|y_s = -1|, |y_s = +1|)$. $Q$ is a truncation parameter to control how rigidly we enforce the consistency within each supervoxel. A higher $Q$ should be set for the supervoxel with more confidence. The penalty $\lambda_{max}(s)$ indicates that the less uniform supervoxel should have less penalty for label inconsistencies. Following~\cite{jain2014supervoxel}, we set it as $\lambda_{max}(s) = |y_s|\exp(-\beta_s \sigma_s)$, where $\sigma_s$ is the RGB variance in the supervoxel $s$ and $\beta_s$ is set as the inverse of the mean of the variances of all supervoxels.

The energy function defined in Eqn.~(\ref{enegy}) can be efficiently minimized using the expansion algorithm~\cite{kohli2009robust}. We set the parameter $Q = 0.1|y_s|$ for all the videos. The optimal label assignments corresponding to the minimum energy yield our desired video-context derived human masks. The mask of each key image $I_k$ is denoted as $l^t_k$. The video-context derived human masks can be utilized to update the CNN. The labeling quality $\omega_k^t$ of each human mask is estimated as the mean of the predicted probabilities on the spatial superpixels which are assigned to be human. { To ensure the data diversity and reduce the effect of noisy labels during training, only up to $5$ key frames with the highest labeling qualities are selected upon the initial candidates.} Different key frames may be selected during the different iterations of the learning process. We also select some negative images (i.e., no human appears) randomly from the frames in which no human instance is detected. The proposed framework can then iteratively refine the human masks by re-estimating the unary potentials based on the improved CNN. {Although only the spatial superpixel nodes with high confident detection results are assigned with high possibilities, the strong spatial and motion coherence constraints, which are incorporated by the pair-wise and higher-order potentials, can effectively facilitate mining more diverse instances with variant poses, views and background clutters.}

\vspace{-2mm}
\section{Experiments}
\textbf{Dataset.} The proposed framework is evaluated on the PASCAL VOC 2012 segmentation benchmark~\cite{voc}. {The performance is measured in terms of pixel intersection-over-union (IoU) on the $person$ class.} The segmentation part of the original VOC 2012 dataset contains 1,464 \emph{train}, 1,449 \emph{val}, and 1,456 \emph{test} images, respectively. Our framework can be boosted by only using the weakly labeled YouTube videos and an imperfect human detector. In total, $160,000$ video-context derived human masks are produced from about $20,000$ video shots, in which about $1/3$ of the images contain no human instances. We also report the results of our variants using the extra 10,582 bounding box annotations and mask annotations, provided by~\cite{hariharan2011semantic}. Extensive evaluation of the proposed method is primarily conducted on the PASCAL VOC 2012 \emph{val} set and we also report the performance on the testing set to compare with the state-of-the-arts by submitting the results to the official PASCAL VOC 2012 server.

\begin{table}\setlength{\tabcolsep}{1pt}
	\caption{Comparison of our models with video-context guided inference (``Ours"), image-based segmentation (``Ours (image-based)"), the version without using pairwise potentials (``Ours (w/o pair)") and the version without using higher-order potentials (``Ours (w/o higher)") in terms of IoU (\%) on PASCAL VOC 2012 validation set. }
	\vspace{-3mm}
	\centering
	\label{tab:learning}\begin{tabular}{|c|c|c|c|c|cccccc|cccccccccc|c}
		\toprule
		Iteration & Ours (image-based) & Ours (w/o pair) & Ours (w/o higher)& Ours\\
		\hline
		1 & 23.4 & 26.2 & 25.9 & 28.1\\
		2 & 30.6 & 40.4 & 42.7 & 48.7\\ 
		3 & 47.8 & 51.5 & 53.4 & 57.2\\
		4 & 58.5 & 62.8 & 63.1 & 65.0\\
		5 & 62.3 & 68.3 & 69.7 & 73.6\\ 
		6 & 68.0 & 74.9 & 73.6 & 77.2\\
		7 & 69.6 & 76.8 & 77.7 & 79.4\\ 
		8 & 70.1 & 79.6 & 79.1 & 80.7\\
		9 & 72.3 & 80.2 & 80.6 & 81.6\\
		\hline
		\textbf{10} & \textbf{72.9} & \textbf{80.5} & \textbf{80.9} & \textbf{81.8}\\ 
		\bottomrule
	\end{tabular}
	\vspace{-6mm}
\end{table}


\textbf{Training Strategies.} {We use the weakly-supervised object detector proposed in~\cite{babylearning} for detecting human individuals, leading to the imperfect detection results. \cite{babylearning} proposed a weakly-supervised learning framework for training object detectors with weakly labeled YouTube videos, where only two annotated bounding boxes for the person label are used to initialize the object detectors, and then massive YouTube videos are used to enhance the object detectors.  We borrow their trained object detectors in this work. Since their pre-trained detectors can output the bounding boxes and confidences of all object classes, we only the detection results for the person category and ignore other results.} The segmentation network is initialized by the publicly released VGG-16 model~\cite{vgg}, which is pre-trained on the ImageNet classification dataset~\cite{imagenet}.  This model is also used by all competitors~\cite{long2014fully}~\cite{hariharan2014hypercolumns}~\cite{dai2014convolutional}~\cite{DeepLabCRF}~\cite{wcrf}~\cite{BoxSup}~\cite{mostajabi2014feedforward}~\cite{CRF-RNN}. We run $10$ iterations for training the DCNN and refining the video-context derived human masks. For each iteration, we use a mini-batch size of 20 for the SGD training. The initial learning rate is set as $0.001$ and divided by 10 after every 20 epochs. The network training is performed for about 60 epochs. In each iteration of our learning process, fine-tuning the network with the refined video-context derived human masks takes about two days on a NVIDIA Tesla K40 GPU. It takes about 2 seconds for testing an image.

\vspace{-2mm}
\subsection{Evaluation of Our Learning Framework}
\vspace{-1mm}

Table~\ref{tab:learning} reports the comparison results of the video-context guided inference and the image-based segmentation in different iterations. The version using image-based segmentation, i.e., ``Ours (image-based segmentation)'', is achieved by only minimizing the unary potential of each node based on the extracted supervoxels in Eqn.~(\ref{enegy}). In this case, the appearance consistency and temporal continuity for the assignments of nodes would not be utilized to generate the human masks.  In terms of our full version (``Ours"), only $28.1\%$ of IoU is obtained by only using the YouTube videos in the beginning. After 10 iterations, we achieve a substantial increase, i.e., obtaining $81.8\%$ of IoU. {The proposed framework is performed for 10 iterations because only slight increase (i.e., $0.2\%$) is observed after 10 iterations.} The increases in IoU are very large in the early iterations (e.g., over $20.6\%$ of IoU after the second iteration), since most of the easy human instances can be recognized and segmented out by the updated network. After the network is gradually improved and the video-context derived human masks are refined, more diverse instances can be collected, which leads to better network capability. Large performance gap in IoU can be observed by comparing ``Ours (image-based segmentation)'' with ``Ours", e.g., $8.9\%$ drop in IoU after 10 iterations. This significant inferiority demonstrates the effectiveness of using the video-context derived inference. {The effectiveness of using pairwise and higher-order potentials can be demonstrated by comparing ``Ours (w/o pair)" and ``Ours (`w/o higher)" with ``Ours", respectively.} 

\begin{table}\setlength{\tabcolsep}{2.6pt}
	\caption{Comparison of different network training strategies in terms of IoU (\%) on PASCAL VOC 2012 validation set.}
	\vspace{-2mm}
	\centering
	\label{tab:network}\begin{tabular}{c|c|c|cccccccc|cccccccccc|c}
		\toprule
		method & sample-weighted loss & negative images &  IoU\\
		\hline
		& No & No & 70.3\\
		& Yes &  No & 74.9\\
		Ours & No &  Yes & 76.8\\ 
		& Yes &  Yes & \textbf{81.8}\\ 
		\bottomrule
	\end{tabular}
	\vspace{-7mm}
\end{table}

\vspace{-3mm}
\subsection{Comparisons of Supervision Strategies}

Table~\ref{tab:final} shows the comparison results of using different strategies of supervision. Training with 160k video-context derived human masks, our method can yield a score of $81.5\%$. {We also report two results by using the extra 10,582 training images with bounding box annotations and mask annotations,  respectively, provided by~\cite{hariharan2011semantic}, which provides the pixel-wise annotations for all 10,582 training images on the 20 object classes.} The bounding box annotations and mask annotations are only used as the extra data to train the network in the last iteration. In terms of using bounding box annotations, we select the region proposals which have largest overlaps with the ground-truth boxes, and then use them as the human masks for training. To combine with the video-context derived human masks for training, we set the labeling quality of the proposal mask from bounding box annotation or mask annotation as $1$, and then fine-tune the network based on the combined set. By using the extra annotated bounding boxes, only a slight $0.4\%$ increase in IoU is obtained. This insignificant change may be because the number (10k) of bounding boxes is small compared to our large number (160k) of video-context derived human masks. When replacing the box annotations with mask annotations,  significant increase can be observed, i.e., $83.0\%$ \emph{vs.} $81.5\%$. This means that carefully annotated masks contain more local detail information or difficult training samples (e.g., extremely small instances or heavily occluded instances) that may be lost within the generated video-context derived human masks.

\begin{table*}\setlength{\tabcolsep}{0.6pt}
	\caption{Results on the PASCAL VOC 2012 test set. In the ``supervision" column, ``mask" means all training samples are with segmentation mask annotations, ``very-weakly" means only the weakly labeled videos and an imperfect human detector are used, ``weakly" means only object class or bounding box annotations are used, and ``semi" means mixture of annotations (e.g., some mask annotations are used).}
	\vspace{-3mm}
	\centering
	\label{tab:final}\begin{tabular}{|c|c|c|c |c|c|c|c|c|c|c|c|ccccccccc|c}
		\toprule
		method & supervision & {prediction type} & mask & box & {auto mask} & auto box & training data & {IoU on person} & {IoU on car}\\
		\hline
		Hyper-SDS~\cite{hariharan2014hypercolumns} &  &  multi-class &  & & & & & 72.9 & 71.9\\
		CFM~\cite{dai2014convolutional} &  & multi-class & & & &  & & 67.5 & 69.2\\ 
		FCN~\cite{long2014fully} &  & multi-class  & & & & & & 73.9 & 74.7\\
		TTI~\cite{mostajabi2014feedforward} & mask & multi-class &  10k & - & - & - & VOC &  68.8 & 74.0\\
		DeepLab-CRF~\cite{DeepLabCRF} &  & multi-class & & & & & & 77.6 & 75.5\\
		DeepLab-CRF-person~\cite{DeepLabCRF} &  & 2-class & & & & & & 76.7 & 74.3\\
		\hline 
		& weakly (object class) & multi-class & - & 10k & - & - &  VOC & 28.2 & 44.9 \\
		& weakly (box)  & multi-class  &  - & 10k & - & - & VOC & 58.2 & 66.5 \\
		WSSL~\cite{wcrf} & semi & multi-class &  1.4k & 9k & - & - & VOC & 76.0 & 76.2\\
		& semi & multi-class  &  2.9k & 8.5k & - & - & VOC & 76.9 & 76.9\\
		& mask & multi-class  &  133k & - & - & - & VOC+COCO & 81.6 & 81.0\\
		\hline
		& weakly (box) & multi-class &  - & {-} & - & - & VOC & 75.1 & 75.0\\
		BoxSup~\cite{BoxSup} & semi & multi-class &  1.4k & 9k & - & - & VOC & 76.9 & 74.6\\
		& semi & multi-class  &  10k & 123k & - & - & VOC+COCO & 81.3 & \textbf{78.5}\\
		\hline
		CRF-RNN~\cite{CRF-RNN}  & mask & multi-class & 133k & - & -& - & VOC+COCO & 81.1 & 76.3\\
		\hline 
		& very-weakly (detector) & 2-class &  - & 2 & 21k & 20k & YouTube & 81.5 & 77.5\\
		Ours & weakly (box) & 2-class & - & 10k & 21k& 20k & VOC+YouTube  & 81.9 & 77.7\\
		& semi &  2-class & 10k & 2 & 21k & 20k &  VOC+YouTube  & \textbf{83.0} & 78.3\\
		\bottomrule
		
	\end{tabular}
	\vspace{-2mm}
\end{table*}

\begin{figure*}[htbp]
	\begin{center}
		\includegraphics[scale=0.27]{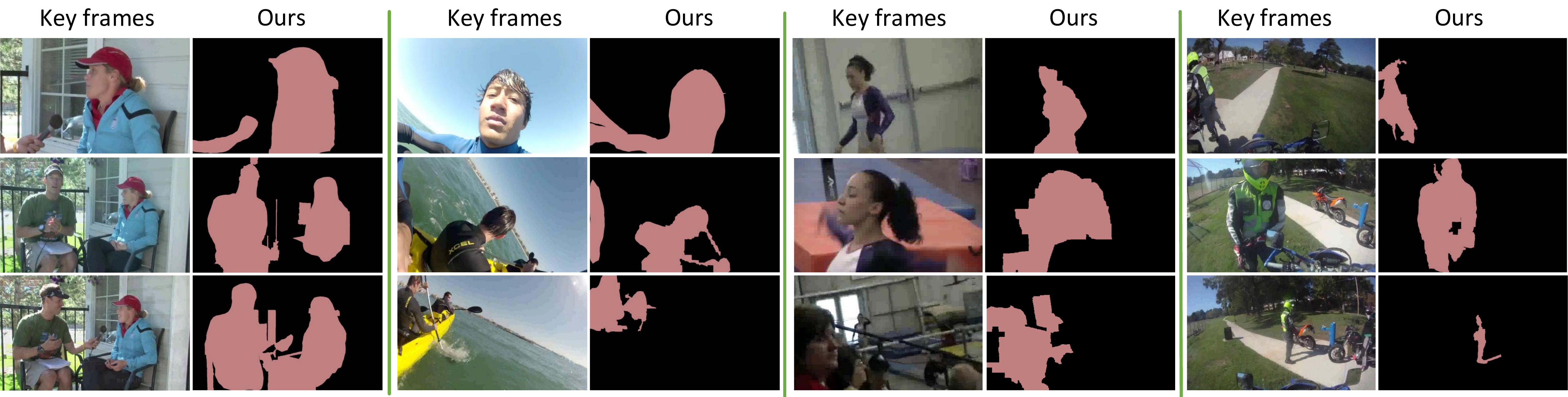}
		\vspace{-1mm}
		\caption{{Exemplars of the generated video-context derived human masks. We display four videos with various background, views and appearances. 
			}}
			\label{fig:videosegmentation}
		\end{center}
		\vspace{-6mm}
	\end{figure*}

\vspace{-3mm}
\subsection{Comparisons of Network Training Strategies}
\vspace{-1mm}

In Table~\ref{tab:network}, we evaluate different network training strategies by using the video-context derived human masks with possibly noisy labels. For the version without using the sample-weighted loss, all video-context derived human masks are treated as contributing equally to training the whole network. In this case, we observe that about $4.4\%$ drop takes place in IoU, compared with our full training strategy. We also validate the effectiveness of using more negative frames collected from raw videos to train the network. About $6.9\%$ decrease in IoU can be observed when comparing the version without using more negative frames with our full version. 

	\begin{figure*}[htbp]
		\begin{center}
			\includegraphics[scale=0.14]{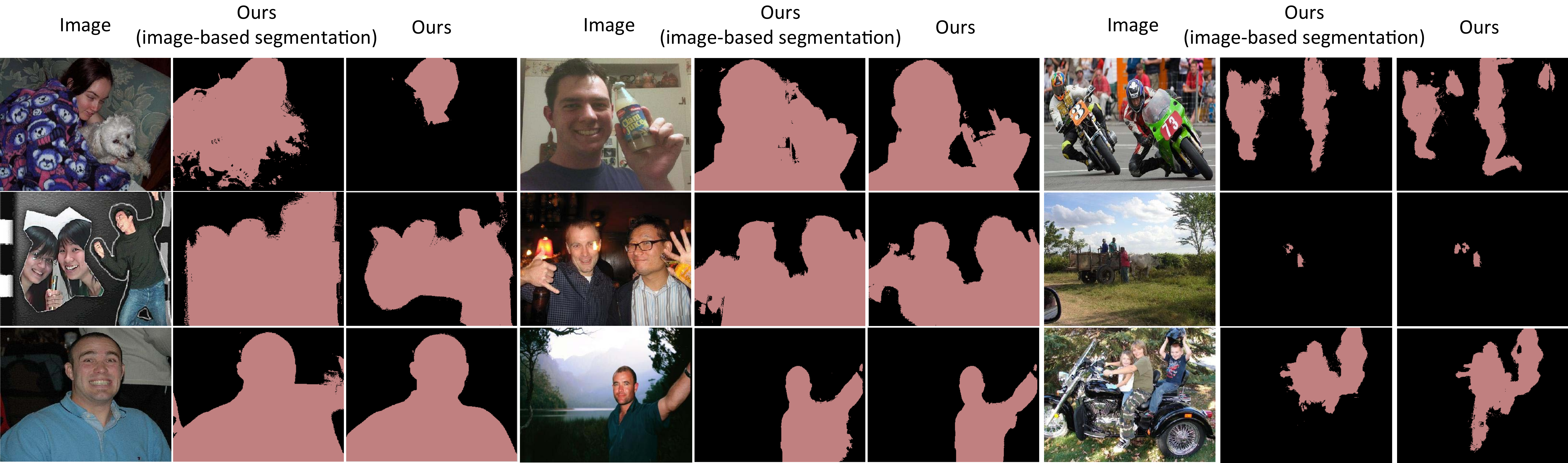}
			\vspace{-2mm}
			\caption{{Example human segmentation results on PASCAL VOC 2012 validation using our method. For each image, we show the results of our version and its variant using image-based segmentation.
				}}
				\label{fig:parsingresults}
			\end{center}
			\vspace{-6mm}
		\end{figure*}

\vspace{-3mm}
{\subsection{Comparisons of Ways of Using videos}
Here we have adopted three simple strategies to evaluate the usage of the video information for boosting the segmentation network. First, we test the performance of directly using the frames of all videos with the image-level ``person'' label as training data. The segmentation networks are thus trained by using the image-level supervision, similar to~\cite{wcrf}, resulting in $34.2\%$ in terms of IoU on the person label, which is better than $28.2\%$ of~\cite{wcrf}. This verifies the large-scale frames in videos can help train better segmentation networks than using the limited number of images. Second, since many frames may not contain any person instances, we further evaluate whether the EM procedure can facilitate improving the capability of the segmentation network. Specifically, ten iterations are performed to progressively reduce the effect of noisy frames. After each iteration, we use the currently trained segmentation network to test all frames, and the frames predicted as containing less than $10\%$ foreground pixels are eliminated during the training in next iteration. We find that employing such EM procedure can obtain $52.8\%$ IoU on person label. Third, we also test the result of using the image co-localization method~\cite{TangCVPR14} to discover the bounding boxes of human instances. The segmentation network can thus be trained with bounding box supervisions of these mined instances. Its final result in terms of IoU on person category is 40.6\%, which is worse than the second strategy. These results of using different strategies further justify the effectiveness of the proposed procedure. 
}

\vspace{-3mm}
\subsection{Comparisons with State-of-the-art Methods}

In Table~\ref{tab:final}, we compare our method with the state-of-the-art methods, including Hyper-SDS~\cite{hariharan2014hypercolumns}, CFM~\cite{dai2014convolutional}, FCN~\cite{long2014fully}, TTI~\cite{mostajabi2014feedforward}, DeepLab-CRF~\cite{DeepLabCRF}, CRF-RNN~\cite{CRF-RNN}, WSSL~\cite{wcrf} and BoxSup~\cite{BoxSup}, on the Pascal VOC 2012 testing set. All these methods use the pre-trained VGG model~\cite{vgg} to initialize the network parameters. Among all of these competitors, WSSL~\cite{wcrf} and BoxSup~\cite{BoxSup} use  different supervision strategies (e.g., object class, bounding box or mask annotations) to train the network. We use the same network setting as in~\cite{wcrf} and~\cite{BoxSup} for fair comparison. We also test the DeepLab-CRF~\cite{DeepLabCRF} method on the 2-class segmentation task (``DeepLab-CRF-person"), i.e., only \emph{person} and \emph{background} classes predicted for each pixel, which is the same setting as used in this paper. The $0.9\%$ decrease in IoU compared with DeepLab-CRF~\cite{DeepLabCRF} may be because the contextual information from the other object classes is not utilized during training the 2-class network. 
Our method that only uses the weakly labeled videos and an imperfect human detector achieves $81.5\%$ in IoU, which is better than the previous fully supervised methods by a considerable margin, e.g., $77.6\%$ of DeepLab-CRF~\cite{DeepLabCRF} and $73.9\%$ of FCN~\cite{long2014fully} on VOC 2012. Remarkably, all of them use all the 10k annotated masks on VOC 2012.

Moreover, we compare our results with other semi-supervised methods~\cite{wcrf}~\cite{BoxSup}. The proposed method is significantly superior over the previous method~\cite{wcrf} which is supervised with object class annotations, i.e., $81.5\%$ \emph{vs.} $30.3\%$. The proposed method achieves $6.4\%$ and $9.1\%$ gain, compared with the methods using bounding box annotations on the VOC dataset, i.e.,  BoxSup~\cite{BoxSup} and WSSL~\cite{wcrf}, respectively. The superiority over WSSL~\cite{wcrf} and BoxSup~\cite{BoxSup} can also be observed when comparing with their semi-supervised variants, i.e., replacing about $86\%$ bounding box annotations with mask annotations on the VOC 2012 dataset. In addition, these previous methods reported the results after augmenting the training data by the large-scale Microsoft COCO dataset~\cite{lin2014microsoft}. The 123,287 images with available mask annotations are provided on COCO. Our results by only using weakly labeled videos and an imperfect human detector are comparable with the fully supervised baselines~\cite{wcrf}~\cite{CRF-RNN} using extensive extra 123k COCO images, e.g., $81.5\%$ of our method \emph{vs.} $81.1\%$ of the CRF-RNN~\cite{CRF-RNN}. It is slightly better than $81.3\%$ of the BoxSup~\cite{BoxSup} using 123k annotated bounding boxes. 

Our semi-supervised variant using the 10k mask annotations on VOC dataset achieves the IoU score of $83.0\%$, which is higher than the performance of all the previous human segmentation methods. 
Most recently, unpublished results in~\cite{piecewise15} reached $82.7\%$ in IoU by using all 10k pixel-wise annotations. This method~\cite{piecewise15} utilized the piecewise training of CRFs instead of the simple fully-connected CRF used by our solution and other state-of-the-arts~\cite{DeepLabCRF}~\cite{wcrf}~\cite{BoxSup}. 

\section{Result Visualization}
We show the video-context derived human masks in the videos obtained by our method in Figure~\ref{fig:videosegmentation}. All masks are generated in the last iteration of the learning process. Although the YouTube videos are often with  low resolution, diverse view points and heavy background clutters, our method can successfully segment out the human instances with different scales or occlusions. In Figure~\ref{fig:parsingresults}, we show the results of our method and its variant using image-based segmentation on the VOC 2012 validation dataset. 

\section{ Conclusion and Future Work}

{In this paper, we present a very-weakly supervised learning framework that learns to segment human by watching YouTube videos along with an imperfect human detector.} In turn, the updated network can help generate more precise video-context derived human masks. This process iterates to gradually improve the network. In future work, we plan to extend our framework to generic semantic segmentation.  

\section*{Acknowledgement}
This work was in part supported by State Key Development Program under Grant NO. 2016YFB1001000 and sponsored by CCF-Tencent Open Fund.

{\small
\bibliographystyle{ieee}
\bibliography{egbib}
}
\end{document}